\documentclass[11pt]{article}

\usepackage[final]{acl}

\usepackage{times}
\usepackage{latexsym}

\usepackage[T1]{fontenc}

\usepackage[utf8]{inputenc}
\usepackage{CJKutf8}
\usepackage{microtype}
\usepackage{pifont}
\newcommand{\cmark}{\ding{51}}
\usepackage{inconsolata}
\usepackage{amsmath}
\usepackage{graphicx}
\usepackage{array}

\renewcommand{\arraystretch}{1.1} 
\usepackage[table]{xcolor}
\usepackage[utf8]{inputenc}
\usepackage{booktabs}
\usepackage{multirow}
\usepackage{graphicx}
\usepackage{subcaption}
\usepackage{amssymb} 
\usepackage[normalem]{ulem}    
\usepackage{xcolor}

\title{SignFLIP: A Unified Model for Sign Language Translation and Generation via Stage-wise Alignment at Scale}

\author{
  Zhaoyi An\textsuperscript{1}\thanks{These authors contributed equally.}
  \quad
  Sihan Tan\textsuperscript{1}\footnotemark[1]
  \quad
  Youngbae Hwang\textsuperscript{2}
  \quad
  Kazuhiro Nakadai\textsuperscript{1}
  \quad
  Rei Kawakami\textsuperscript{1} \\
  \textsuperscript{1}Institute of Science Tokyo
  \qquad
  \textsuperscript{2}Chungbuk National University \\
  \texttt{an.z.b041@m.isct.ac.jp}
}

\begin{document}
\maketitle
\begin{abstract}


Sign language translation and generation share the goal of bidirectional alignment between text and sign representations. However, existing approaches either treat them as isolated tasks or are only verified on limited datasets, limiting effective modeling between modalities. In this paper, we propose SignFLIP, 
a unified LLM-centered framework for translation and generation.
To enable bidirectional mapping between text and sign, SignFLIP adopts a symmetric architecture together with a stage-wise training strategy built on large-scale data. The shared sign--text representation is progressively refined: pre-alignment facilitates subsequent SLT, while the SLT-adapted representation further benefits SLG.
Extensive experiments on multiple benchmarks show that SignFLIP shows competitive performance compared with task-specific models on both translation and generation tasks, as well as strong transferability to sign language recognition.




\end{abstract}


\section{Introduction}


\begin{figure}[t]
    \centering
  \includegraphics[width=\columnwidth]{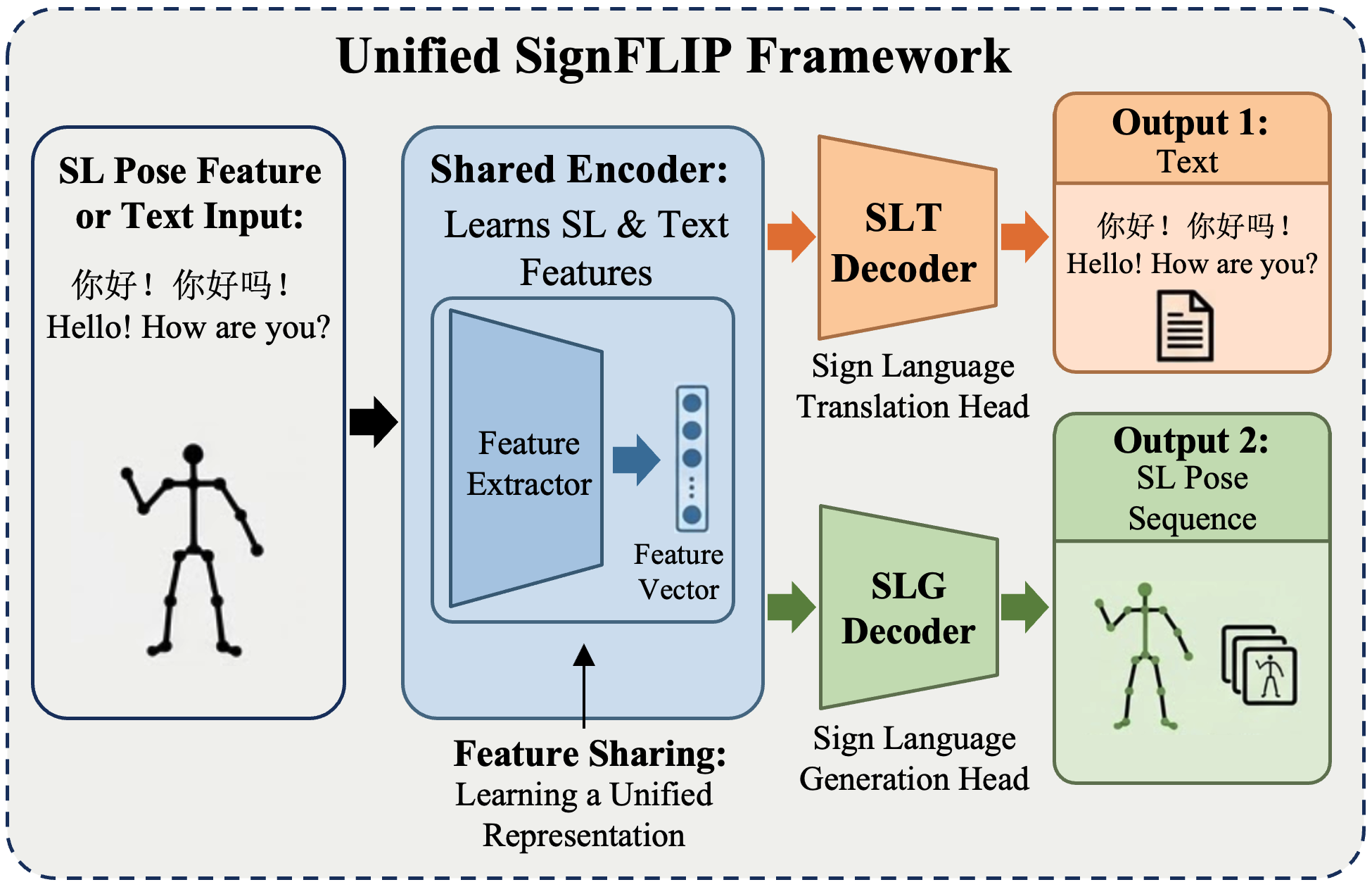}
  \caption{
  Framework of SignFLIP, a unified model for SLT and SLG. A shared encoder learns joint representations from sign features and text inputs, which are then used by task-specific decoders to generate spoken text and sign poses. 
  SignFLIP provides a unified framework for SLT and SLG while achieving competitive performance on both tasks.
  }
  \label{fig:introduction}
\end{figure}


Constructing unified representations across modalities is central to multimodal alignment and plays a crucial role in enabling downstream models, which also holds for sign language processing (SLP). However, in contrast to the rapid progress in broader multimodal learning, unified representation for sign languages (SLs) and texts remains relatively underexplored~\cite{hwang2025spatio}. Most existing studies treat sign-to-text mapping, \textit{e.g.}, sign language translation (SLT)~\cite{8578910,tan-etal-2025-improvement} and sign language recognition (SLR)~\cite{,NEURIPS2022_6cd3ac24,Zuo_2022_CVPR}, and text-to-sign mapping, \textit{e.g.}, sign language generation (SLG)~\cite{yin2024t2s,10.1007/978-3-031-72967-6_3}, as individual problems. This separation is partly driven by the distinct modeling requirements of the two directions, where task-specific architectures are often designed to optimize performance for each individual task~\cite{moryossef2021slp}. Such task-specific designs, however, lead to repeated modeling effort and cannot fully benefit from the inherent duality in SLT and SLG.

Meanwhile, advances in large language models (LLMs)~\cite{brown2020language, yang2025qwen3} have reshaped a wide range of tasks, opening new possibilities for SLT and SLG. 
Trained on massive general-purpose corpora, these models encode rich knowledge and strong reasoning capabilities. 
Although SLs remain underrepresented in such pretraining corpora, recent studies have adapted general-purpose models to the SL domain through fine-tuning~\cite{wong2024signgpt,10.5555/3737916.3741537,hwang-etal-2025-efficient} and prompt engineering~\cite{an2025teach}, improving performance while reducing reliance on heavily task-specific designs~\cite{li2025uni}. 
These advances suggest that pretrained general-purpose models may serve as a shared foundation for SLP, naturally raising the question: \textit{Can we leverage a general LLM to build unified model that serve both SLT and SLG? If so, how?}


In this paper, we propose SignFLIP which adopts a unified architecture (\S~\ref{sec:model archi}) together with stage-wise training at scale using 600K$\sim$700K samples (\S~\ref{subsec:training}). The stages progressively refine a shared sign–text representation: pre-alignment facilitates SLT, while the SLT-adapted representation further benefits SLG. Extensive experiments demonstrate that SignFLIP consistently achieves competitive results on both SLT and SLG across multiple main benchmarks, including CSL-Daily~\cite{zhou2021improving} and How2Sign~\cite{Duarte_CVPR2021}. We further evaluate the transferability of the learned representations on downstream SLR tasks, including isolated and continuous SLR. Our main contributions are as follows:

\begin{itemize}
    \item We propose SignFLIP, a framework for learning
    sign--text mappings within the latent space of a general LLM through large-scale training.
    \item We introduce a stage-wise alignment strategy that progressively refines shared sign–text representations, enabling knowledge learned in earlier stages to benefit subsequent tasks.
    \item We show that SignFLIP achieves competitive performance on both SLT and SLG, while learning transferable representations that further benefit downstream isolated and continuous SLR.
\end{itemize}

\begin{figure*}[t]
  \centering
  \includegraphics[width=\textwidth]{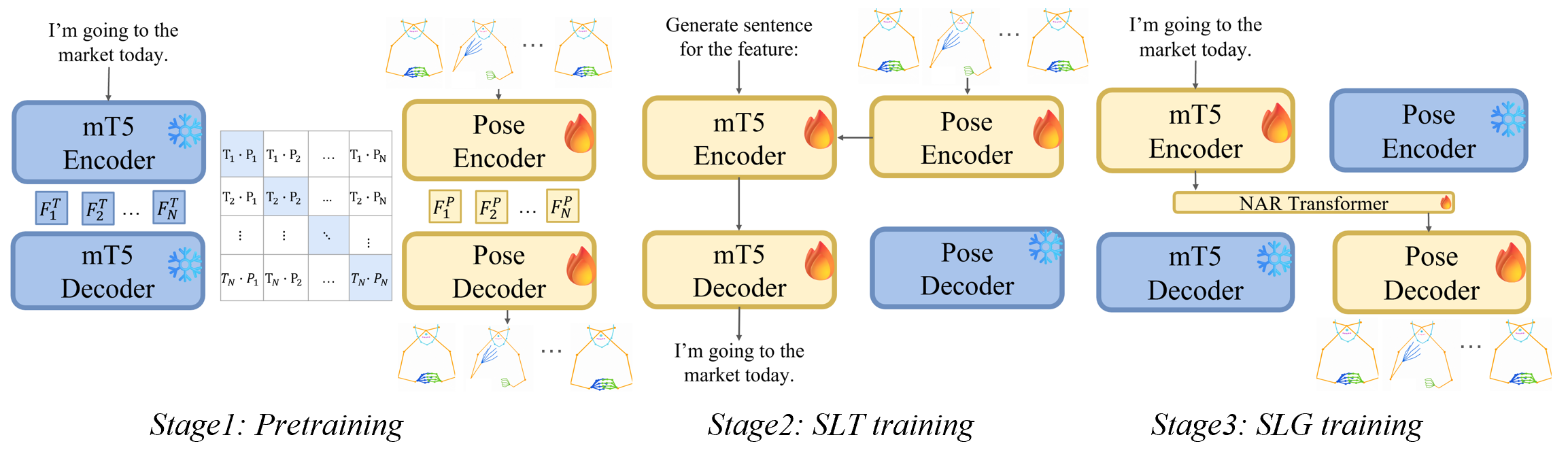}
  \caption{SignFLIP is trained in three stages. Stage 1 performs pretraining to align sign and text embeddings through reconstruction and contrastive learning. Stages 2 and 3 focus on downstream SLT and SLG tasks, respectively. In particular, Stage 1 unfreezes the pose encoder and pose decoder. Stage 2 trains the pose encoder together with mT5 for text generation. Stage 3 then optimizes the mT5 encoder, projector, and pose decoder for sign generation. }
  \label{fig:pipeline}
\end{figure*}
\section{Related Work}

\subsection{Sign Language Pretraining}


In SLP, pretraining has primarily targeted SLU tasks. 
Earlier work, constrained by limited data, often relied on self-supervised pretraining (\textit{e.g.}, SignBERT and SignBERT+~\cite{10109128,Hu_2021_ICCV}) or weakly supervised pretraining~\cite{Zhou_2023_ICCV} on relatively small SL corpora. 
More recently, SLP has increasingly adopted a paradigm similar to that in speech and vision--language modeling: (pre-) training modern models on large (but often noisy) out-of-domain datasets, followed by fine-tuning on smaller in-domain datasets. 
For instance, ~\citet{jiang-etal-2024-signclip} performed contrastive vision--language pretraining on a multilingual SL dictionary, \textit{Spreadthesign}\footnote{\url{https://www.spreadthesign.com/}. The use of the data is under a license granted by Spreadthesign.}, and achieved competitive performance on in-domain sign retrieval as well as out-of-domain isolated SLR tasks. 
Inspired by self-supervised speech pretraining, ~\citet{gueuwou-etal-2025-shubert} proposed ShuBERT for American Sign Language (ASL).~\citet{10.5555/3737916.3741537} scaled SLT through cross-lingual and cross-modal transfer.
Concurrently, Uni-Sign~\cite{li2025uni}, a supervised pretrained model trained on large ASL and Chinese Sign Language (CSL) corpora, has been applied to a range of SLU tasks.
Nevertheless, these advances predominantly benefit SLU, and pretraining that directly supports SLT and SLG remains underexplored. 
This gap motivates increasing interest in bridging SLU and SLG within a unified framework.


\subsection{Sign Language Translation}
\paragraph{Gloss-based} method introduces gloss as intermediate supervision to facilitate downstream text generation\footnote{Gloss is another written form of text in sign language order}. SLRT~\cite{camgoz2020sign} was among the first to introduce a Transformer-based encoder–decoder framework for SLT, incorporating gloss-level supervision into the encoder with a CTC loss, while SLTUNET~\cite{zhang2023sltunet}, scaledSLT~\cite{10.5555/3737916.3741537}, and MMTLB~\cite{9879103} explore the use of large-scale external text corpora and pretrained language models to enhance translation quality. While gloss-based approaches have shown promising results, they are fundamentally constrained by the limited availability of paired sign–gloss–text data, which restricts their scalability across multiple downstream tasks.
\paragraph{Gloss-free.}
In contrast, gloss-free SLT directly converts SLs into spoken texts, but it underperformed cascading SLT due to the challenging sign-text alignment. Recent work further advanced gloss-free SLT. YouTube-ASL~\cite{10.5555/3666122.3667386} uses a language modeling objective for large-scale pretraining, showing the potential of generative pretraining and the value of scaling up datasets.~\citet{jang2025lost} leveraged additional contextual cues together with the SL videos to improve translation. 
SignFLIP focuses on gloss-free SLT, pretrained on large-scale data. 

\subsection{Sign Language Generation}
Similar to the translation task, SLG studies are also categorized into gloss-based and gloss-free models. An early gloss-based model is Progressive Transformer~\cite{saunders2020progressive}, which adopts transformer architecture to generate gloss from text, then pose from gloss.~\citet{huang2022dualsign} present a semi-supervised two-stage SLG framework that reduces gloss annotation requirements but is still constrained by the initial and pseudo gloss quality.
Recently, more researchers have shifted toward gloss-free methods that require no extra supervision~\cite{10.1145/3716553.3750765}. These approaches have begun to achieve results comparable to gloss-based ones, particularly with the integration of LLMs \cite{an2025teach,Zuo_2025_ICCV}.

 Despite these advancements and the integration of LLMs, few studies have explored how to jointly design translation and generation modules. USLNet~\cite{guo-etal-2024-unsupervised}, an unsupervised model for SLT and SLG, achieves a translation performance of BLEU 6.30 and a generation performance of FVD 390.5 on OpenASL. UniGloR~\cite{hwang2025spatio} explores spatio-temporal features as an alternative to glosses for SLT and SLG. Along this direction, SignFLIP advances unified SLT and SLG by introducing an LLM-centered architecture trained on large datasets, enabling bidirectional sign--text modeling and systematic evaluation across multiple datasets.
 

\section{SignFLIP}
SignFLIP provides a unified framework for SLT and SLG, with its overall three-stage training pipeline illustrated in Figure~\ref{fig:pipeline}. In Stage 1, the model is pretrained to align sign and text representations while reconstructing sign poses from masked pose sequences. Stage 2 adapts the model to the downstream SLT task, and Stage 3 further trains it for sign pose generation.
\subsection{Model Architecture}
\label{sec:model archi}
Overall, SignFLIP consists of a sign pose encoder, 
an mT5 encoder that processes both text inputs and pose-derived features,
a pose decoder that mirrors the architecture of the pose encoder, and an mT5 decoder for text generation.
\paragraph{SignFLIP for translation} comprises a sign pose encoder and an mT5 
encoder-decoder 
model~\cite{xue-etal-2021-mt5}. Given 133 keypoints, we selectively retain 69 of them, including 21 keypoints for each hand, 9 for the body, and 18 for the face. The keypoint sequence of the group \(i\) is denoted by \(P_i\), where \(i \in \{h, b, f\}\). Specifically, the keypoint sequence $P_i$ is first encoded by a three-layer spatial GCN, yielding pose features $\mathcal{F}^{P}_{i} \in \mathbb{R}^{L \times N_i \times C}$, where $L$ is the temporal length of the keypoint sequence, $N_i$ is the number of keypoints in group $i$, and $C$ is the feature dimension. These pose features are subsequently projected to 
the hidden dimension of mT5 and are then fed into mT5 for autoregressive text generation.

\paragraph{SignFLIP for generation} involves an mT5 encoder for text processing and a non-autoregressive transformer for feature alignment, as well as a pretrained pose decoder for generating poses from features. Given the input spoken sentence $T_i$ with $U$ tokens, the mT5 encoder processes it into text features $\mathcal{F}^{T}_{i} \in \mathbb{R}^{U_i \times D}$, where $U_i$ denotes the token length of the input sentence and $D$ denotes the hidden dimension of the mT5 encoder. Since the frame length of the pose sequence is generally longer than the length of text tokens, in the non-autoregressive transformer, we first predict the sequence length from text feature input, then employ learnable initial tokens to interact with the text feature via cross-attention. From the generated pose feature, we decode it into the final pose sequence with the pose decoder.

\subsection{Stage 1: Pretraining}
\label{subsec:training}
In Stage 1, we train SignFLIP's pose encoder and pose decoder, which are fully symmetric in architecture: a three-layer spatial GCN. As they can operate directly on human skeletons in non-Euclidean space, preserving the physical topology of SLs.
We optimize their parameters by two objectives: semantic-aware pose reconstruction and sign--text pre-alignment.

\paragraph{Semantic-aware Pose Reconstruction.} 
To ensure the extracted features capture semantic information, we randomly mask $n$ consecutive frames out of a 24-frames window using a learnable mask token. This requires the pose encoder and decoder to not only learn kinematic information but also infer features at the masked positions based on the surrounding context, thereby improving the model's contextual modeling ability and robustness. We chose this masking strategy because, in SLP tasks where inter-frame variations are small, a model can easily recover a single missing frame through simple linear interpolation. By masking several consecutive frames, the model is forced to interact with the context to infer the missing sign information, ensuring that the generated pose features carry semantic meaning. This process can be formulated as:
\begin{align}
    \hat{p}&=\mathcal{D}(\mathcal{E}(p_\text{mask})), \\
    \mathcal{L}_\text{recon}&=\text{SmoothL1}(\hat{p},p),
\end{align}
where $p, p_\text{mask}, \hat{p}$ denote for input pose sequence, masked pose and reconstructed pose respectively, while $\mathcal{D}, \mathcal{E}$ represent the pose encoder and decoder.

\paragraph{Sign--Text Pre-alignment.}
While the reconstruction task embeds semantic information into pose features, they still lie in a different manifold with the LLM-derived text features. We regard the text features from the LLM as anchors, as it has been pretrained on trillion scale data, and adopt contrastive learning to perform alignment at the initial stage and reduces the difficulty 
of downstream tasks.
The overall pretraining loss is defined as:
\begin{align}
    \mathcal{L}_\text{align} &= - \log \frac{\exp(\mathrm{cos}(s_i, t_i)/\tau)}{\sum_{j=1}^{N} \exp(\mathrm{cos}(s_i, t_j)/\tau)}, \\
    \mathcal{L}_\text{Pretrain} &= \mathcal{L}_\text{recon} + \alpha \mathcal{L}_\text{align},
\end{align}
where the $(s_i, t_i)$ denotes the sign-text pair and $\tau$ means the temperature. The values of $\alpha$ and following hyperparameters that we adopt at training are shown in Appendix~\ref{sec:training_recipe}.

\subsection{Training on Downstream Task}

\textbf{Stage 2: SLT training.} In this stage, we unfreeze both the pose encoder and the language model (shown in Figure~\ref{fig:pipeline}, Stage 2). The goal is to maintain the strong translation capability of SignFLIP while allowing the mT5 encoder to adapt to pose features through task-specific training, thereby preparing it for the subsequent generation stage. To this end, we first train SignFLIP on a large-scale corpus with noise and then fine-tune it on the target dataset. Accordingly, we optimize the model with the standard autoregressive text generation objective, defined as follows:
\begin{equation}
\mathcal{L}_{\text{SLT}} 
= - \sum_{u=1}^{U} \log p\left(t_u \mid t_{<u}, \mathcal{F}^{P}_{i}\right),
\end{equation}
where $U$ denotes the length of the target text sequence, $t_u$ is the $u$-th target token, $t_{<u}$ represents all previously generated target tokens before step $u$, and $\mathcal{F}^{P}_{i}$ denotes the pose feature from the pose encoder.
\paragraph{Stage 3: SLG training.}
In contrast to the autoregressive generation common in decoder-only LLMs, we argue that a non-autoregressive (NAR) approach, which produces the entire sequence in a single pass, is better suited for sign language since the model can consider the global information as well as SL's grammar to arrange output. In our SLG framework, the text and prompt are first processed by an mT5 encoder to extract global semantic features. These features are then used to predict the pose sequence length through a MLP. Subsequently, learnable pose tokens with predicted length interact with the text features via cross-attention to generate pose features, which are decoded into the final pose sequence. Leveraging the scale of current sign language datasets, we avoid designing alignment modules with heavy inductive bias, instead allowing the attention mechanism and the LLM to learn the cross-modal mapping directly.

However, we find that the text features generated by the LLM encoder often lack sufficient granularity. As a result, after interacting with these features, tokens struggle to reconstruct the fine-grained details of pose features, such as finger movements, and instead collapse toward the mean value.
To address this, besides the standard Smooth L1 reconstruction loss, we use the pose encoder pretrained on the reconstruction task to supervise the output of the alignment module through feature distillation. This guides the generated features to align with the ideal pose representations. Additionally, we incorporate a hand-specific loss in relative coordinates to increase the penalty for hand errors. The total loss for SLG is formulated as follows:
\begin{align}
    \mathcal{L}_{\text{Distill}} &= (1 - \cos(\hat{F}_i^P, F_i^P)) + \text{L2}(\hat{F}_i^P, F_i^P) \\
    \mathcal{L}_{\text{Hand}} &= \text{SmoothL1}\bigl((\hat{P}^{\text{hand}}_k -\hat{P}^{\text{wrist}}) \nonumber \\
    &\quad - (P^{\text{hand}}_k - P^{\text{wrist}})\bigr) \\
    \mathcal{L}_{\text{SLG}} &= \mathcal{L}_{\text{Recon}} + \beta \cdot \mathcal{L}_{\text{Distill}} + \gamma \cdot \mathcal{L}_{\text{Hand}},
\end{align}

where $\hat{F_i^P}$ and $F_i^P$ denote the pose feature from NAR transformer and pose encoder respectively, $P_k^\text{hand}$ represents the keypoint joints of the hand part and $k \in [1, K]$ is the keypoint id of hands. When training on the SLG task, we set the learning rate of mT5 encoder and pose decoder as $1\%$ of the NAR transformer's lr to prevent feature shift, also to minimize performance degradation for SLT. 

\section{Experiments}
\label{Sec:Experiment}

\subsection{Implementations and Preprocessing}
For large-scale pretraining, we adopt CSL-News~\cite{li2025uni} and YouTubeASL~\cite{10.5555/3666122.3667386} for Chinese Sign Language and American Sign Language, respectively. For pose extraction, we employ RTMPose-x~\cite{jiang2023rtmposerealtimemultipersonpose}, implemented in MMPose\footnote{\url{https://github.com/open-mmlab/mmpose/tree/main/projects/rtmpose}}, to obtain whole-body keypoints from original RGB videos. We use mT5-Base~\cite{xue-etal-2021-mt5} as the pretrained LLM and conduct all experiments on four NVIDIA H100 GPUs. The training recipes of each stage are shown in Appendix~\ref{sec:training_recipe}.

\begin{table*}[htb]
\centering
\setlength{\fboxsep}{1pt}
\resizebox{0.75\textwidth}{!}{%
\begin{tabular}{lcccccc}
\toprule
\multirow{2}{*}{\textbf{Method}} & \multicolumn{2}{c}{\textbf{Modality}} & \multicolumn{2}{c}{\textbf{Dev}} & \multicolumn{2}{c}{\textbf{Test}} \\
\cmidrule(lr){2-3} \cmidrule(lr){4-5} \cmidrule(lr){6-7}
& \textbf{Pose} & \textbf{RGB} & \textbf{BLEU-4}$\uparrow$ & \textbf{ROUGE} $\uparrow$& \textbf{BLEU-4} $\uparrow$& \textbf{ROUGE} $\uparrow$\\
\midrule

\rowcolor{gray!20}
\multicolumn{7}{c}{\textbf{Gloss-based}} \\
SLRT$^\dagger$~\cite{camgoz2020sign} &        & \cmark & 11.88 & 37.96 & 11.79 & 36.74 \\
MMTLB~\cite{9879103}                 &        & \cmark & 24.42 & 53.38 & 23.92 & 53.25 \\
TS-SLT~\cite{NEURIPS2022_6cd3ac24}  & \cmark & \cmark & \underline{25.76} & \underline{55.10} & \underline{25.79} & \underline{55.72} \\
SLTUNET~\cite{zhang2023sltunet}     &        & \cmark & 23.99 & 53.58 & 25.01 & 54.08 \\

\midrule

\rowcolor{gray!20}
\multicolumn{7}{c}{\textbf{Gloss-free}} \\
SLRT$^\ddagger$~\cite{camgoz2020sign} &        & \cmark & 4.04  & 20.51 & 3.03  & 19.67 \\
GF-SLT~\cite{Zhou_2023_ICCV}          &        & \cmark & 11.07 & 36.70 & 11.00 & 36.44 \\
SignLLM~\cite{Gong2024LLMsAG}         &        & \cmark & 12.23 & 39.18 & 15.75 & 39.91 \\
C$^2$RL~\cite{Chen2024C2RLCA}         &        & \cmark & --    & --    & 21.61 & 48.21 \\
MSLU~\cite{11125948}                  & \cmark &        & 10.27 & 33.13 & 11.42 & 33.80 \\
Sign2GPT~\cite{wong2024signgpt}       &        & \cmark & --    & --    & 22.52 & 48.90 \\
SpaMo~\cite{hwang-etal-2025-efficient}&        & \cmark & --    & --    & 20.55 & 47.46 \\
Uni-Sign~\cite{li2025uni}             & \cmark &  &  25.27&54.34&25.61&54.92\\
Uni-Sign~\cite{li2025uni}             & \cmark & \cmark & 26.25 & 56.03 & 26.36 & 56.51 \\
Geo-Sign~\cite{fish2025geosign}& \cmark & & \colorbox{cyan!18}{\textbf{27.05}} & \colorbox{cyan!18}{\textbf{57.27 }}& \colorbox{cyan!18}{\textbf{27.42}}& \colorbox{cyan!18}{\textbf{57.95}}\\
\midrule
\textit{Unified}\\
\textbf{SignFLIP (Ours)}              & \cmark &        & \colorbox{green!18}{\textbf{25.71}} &\colorbox{green!18}{\textbf{54.67 }}& \colorbox{green!18}{\textbf{26.10 }}& \colorbox{green!18}{\textbf{55.24 }}\\
\bottomrule
\end{tabular}%
}
\caption{SLT results on the CSL-Daily dataset. $\dagger$ and $\ddagger$ denote results reproduced by \citet{zhou2021improving} and \citet{Zhou_2023_ICCV}, respectively. Underlined results indicate the best performance among gloss-based SLT methods, \colorbox{cyan!18}{Blue} and \colorbox{green!18}{Green} denote
the best results of previous methods and ours, respectively. SignFLIP achieves competitive performance among pose-only methods and remains competitive with RGB-based and multimodal approaches.}
\label{tab:slt_csl}
\end{table*}

\begin{table*}[!htb]
\centering
\resizebox{0.7\textwidth}{!}{%
\begin{tabular}{lccccc}
\toprule
\multirow{2}{*}{\textbf{Method}} & \multicolumn{2}{c}{\textbf{Modality}} & \multicolumn{3}{c}{\textbf{Test}} \\
\cmidrule(lr){2-3} \cmidrule(lr){4-6}
& \textbf{Pose} & \textbf{RGB} & \textbf{BLEU-4} $\uparrow$ & \textbf{ROUGE} $\uparrow$& \textbf{BLEURT} $\uparrow$\\
\midrule

\rowcolor{gray!20}
\multicolumn{6}{c}{\textbf{How2Sign}} \\
GloFE-VN~\cite{lin-etal-2023-gloss}         & \cmark &        & 2.2  & 12.6 & 31.7 \\
YouTube-ASL~\cite{10.5555/3666122.3667386}  & \cmark &        & 12.4 & --   & 46.6 \\
MSLU~\cite{11125948}                        & \cmark &        & 2.4  & 17.2 & --   \\
C$^2$RL~\cite{Chen2024C2RLCA}               &        & \cmark & 9.4  & 27.0 & --   \\
FLa-LLM~\cite{chen-etal-2024-factorized}    &        & \cmark & 9.7  & 27.8 & --   \\
SSVP-SLT~\cite{rust-etal-2024-towards}      &        & \cmark &
15.5 &
\colorbox{cyan!18}{\textbf{38.4}} &
49.6 \\
ShuBERT~\cite{gueuwou-etal-2025-shubert}    & \cmark & \cmark &
\colorbox{cyan!18}{\textbf{16.2}} &
-- &
\colorbox{cyan!18}{\textbf{49.9}} \\
Uni-Sign~\cite{li2025uni}  &\cmark&& 14.5&34.3&48.6\\
Uni-Sign~\cite{li2025uni}                   & \cmark & \cmark &
14.9 &
36.0 &
49.4 \\
Geo-Sign~\cite{fish2025geosign} & \cmark & &15.1& 35.4& --\\
\midrule
\textit{Unified}\\
UniGloR~\cite{hwang2025spatio} & \cmark && 2.22 & 12.98 & --\\
\textbf{SignFLIP (Ours)}                    & \cmark &  & \colorbox{green!18}{\textbf{15.1}} &  \colorbox{green!18}{\textbf{36.8}} & \colorbox{green!18}{\textbf{49.8}}\\




\bottomrule
\end{tabular}%
}
\caption{SLT results on How2Sign.  SignFLIP continues to outperform existing pose-only methods on How2Sign dataset.
}
\label{tab:slt_asl}
\end{table*}

\subsection{Datasets and Evaluations}
\paragraph{Datasets.}

We evaluate SignFLIP on CSL-Daily~\cite{zhou2021improving}, OpenASL~\cite{shi-etal-2022-open}, How2Sign~\cite{Duarte_CVPR2021}, and WLASL100~\cite{li2020word} to demonstrate the effectiveness and transferability of our unified modeling framework. 
Dataset statistics are provided in Appendix~\ref{sec:selected_datasets}.

\paragraph{Evaluation metrics.} For SLT, following previous work~\cite{camgoz2020sign,10.5555/3737916.3741537}, we adopt BLEU~\cite{papineni-etal-2002-bleu}, computed with SacreBLEU~\cite{post-2018-call}, and ROUGE-L~\cite{lin-2004-rouge} as evaluation metrics. For ASL datasets, we additionally report BLEURT~\cite{sellam-etal-2020-bleurt} scores using the BLEURT-20 checkpoint. For SLG, at pose level, we report Dynamic Time Warping and Mean Joint Error scores. For the evaluation of semantic accuracy, following previous work~\cite{saunders2020progressive}, we report the back-translation scores of BLEU-4. 

For SLG, we further conducted human evaluations with 2 professional ASL and CSL signers, who were asked to rate the generated sign poses from SignFLIP on a 1--5 scale in terms of both semantic accuracy and kinematic quality. The latter was further assessed along three dimensions: motion smoothness, anatomical plausibility, and hand clarity. Specifically, 15 generated sign samples were provided for evaluation. The full evaluation questionnaire is provided in Appendix~\ref{sec:questionaire}.


\subsection{Comparison with Methods}

\paragraph{Results on SLT.} Using pose data only, SignFLIP achieves a test BLEU-4 of 26.10 and a ROUGE score of 55.24 on CSL-Daily, a test BLEU-4 of 15.1 and a BLEURT score of 49.8 on How2Sign, and a test BLEU-4 of 19.77 and a BLEURT score of 59.35 on OpenASL (Appendix~\ref{sec:openasl}). Tables~\ref{tab:slt_csl} and~\ref{tab:slt_asl} show that, under the pose-only setting, SignFLIP sets a new state of the art on How2Sign while remaining competitive on CSL-Daily. Notably, despite the performance drop caused by the later SLG finetuning stage, it still delivers strong results across all datasets and outperforms several RGB-based or multimodal-based SLT methods, demonstrating the benefits of large-scale pretraining. 

\begin{table}[t]
\centering
\small
\setlength{\tabcolsep}{4pt} 
\renewcommand{\arraystretch}{1.15}
\begin{tabular}{m{0.42\columnwidth} >{\centering\arraybackslash}m{0.14\columnwidth} >{\centering\arraybackslash}m{0.18\columnwidth} >{\centering\arraybackslash}m{0.14\columnwidth}}
\toprule
\textbf{Method} & \textbf{DTW}$\downarrow$ & \textbf{DTW--MJE}$\downarrow$ & \textbf{BLEU-4}$\uparrow$ \\
\midrule

\rowcolor{gray!15}
\multicolumn{4}{c}{\textbf{How2Sign}} \\
Progressive Transformer\\
\cite{saunders2020progressive} & 0.397 & 8.98e--4 & 1.63 \\
Teach Me Sign\\
\cite{an2025teach} & 0.345 & 7.20e--4 & 3.95 \\
\midrule
\textit{Unified}\\
\textbf{SignFLIP (Ours)} & \colorbox{green!18}{\textbf{0.248}} & \colorbox{green!18}{\textbf{5.28e--4}} & \colorbox{green!18}{\textbf{7.10}} \\
\midrule

\rowcolor{gray!15}
\multicolumn{4}{c}{\textbf{CSL-Daily}} \\
Progressive Transformer\\
\cite{saunders2020progressive} & 0.048 & 1.16e--4 & 2.42 \\
Teach Me Sign\\
\cite{an2025teach} & 0.035 & 8.86e--5 & 4.87 \\
\midrule
\textit{Unified}\\
\textbf{SignFLIP (Ours)} & \colorbox{green!18}{\textbf{0.025}} & \colorbox{green!18}{\textbf{6.15e--5}} & \colorbox{green!18}{\textbf{9.75}} \\
\bottomrule
\end{tabular}
\caption{SLG results on the How2Sign and CSL-Daily datasets. Metric scales differ across datasets due to variations in generation difficulty and keypoint ranges.}
\label{tab:slg}
\end{table}
\begin{figure*}[!htb]
\centering
\includegraphics[width=\textwidth]{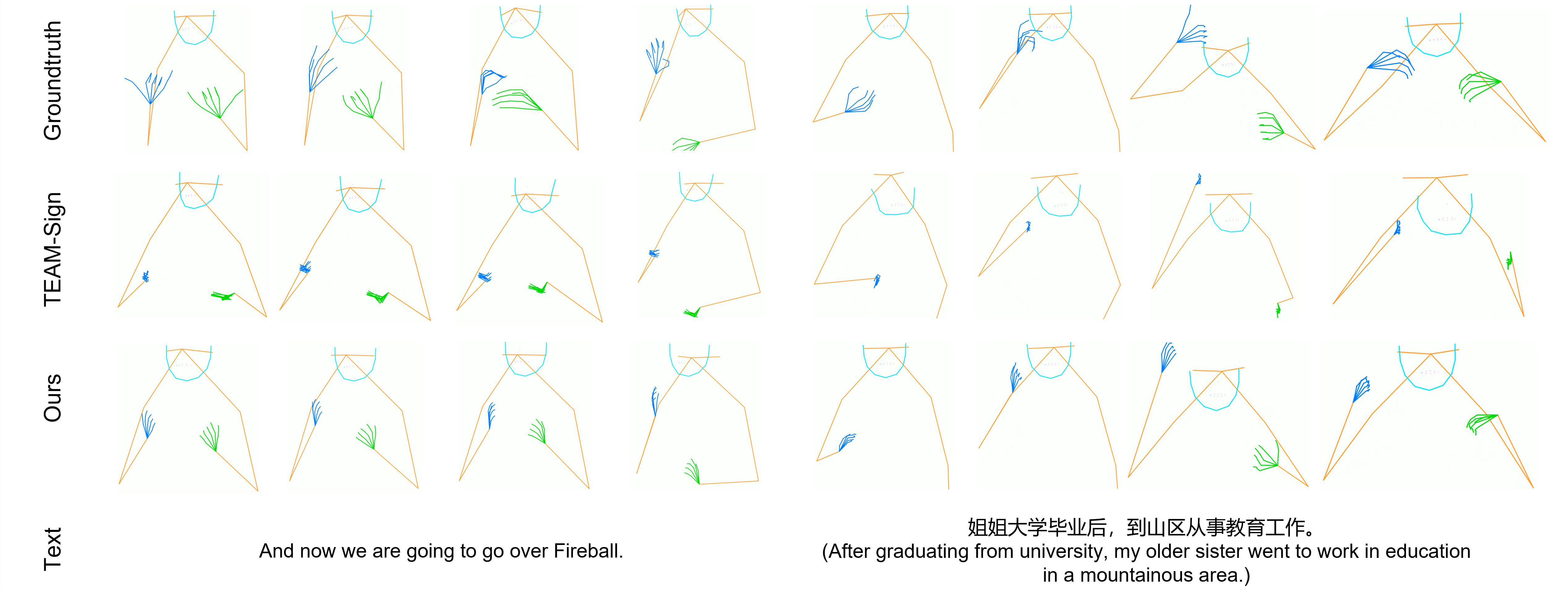} 
\caption{Qualitative results on both How2Sign (left) and CSL-Daily (right) datasets. While the arm movements of the poses generated by our proposed SignFLIP are closer to the Ground Truth, our generated hand motions are also more natural than those from previous methods, by leveraging the feature distillation loss and hand-specific loss.}
\label{fig:slg-qua}
\end{figure*}

\begin{table}[!htb]
    \centering
    \scalebox{0.95}{
    \begin{tabular}{lcc}
    \toprule
    \textbf{Method} & \textbf{How2Sign} & \textbf{CSL-Daily} \\
    \midrule
    \textbf{SignFLIP (Ours)} & 1.69 & 1.94\\
    GT motions  & 3.48 & 3.44 \\
    \bottomrule
    \end{tabular}
    }
    \caption{Human evaluation scores for generated poses on How2Sign and CSL-Daily datasets.}
    \label{tab:human_eval_result}
\end{table}

\paragraph{Results on SLG.} 
Given that SLG previous works generate various modalities of sign language videos, such as 2D poses, 3D poses, and 3D meshes, we select Progressive Transformer~\cite{saunders2020progressive} and Teach Me Sign~\cite{an2025teach} for comparison, as they both produce 2D poses as our model. We retrained these models on the CSL and ASL datasets.
Quantitative results are tabulated in Table \ref{tab:slg}. We observe that the proposed model outperforms previous work at both the kinematic and semantic levels of the generated poses. 
Note that since the scale of DTW and MJE metrics is heavily influenced by the scale of the original keypoint data and normalization, cross-model comparisons can be challenging unless the models are trained on the same data. A qualitative comparison is displayed in Figure~\ref{fig:slg-qua}. The keypoint videos generated by SignFLIP align more closely with the ground truth in terms of general movements, especially arm trajectories. While synthesizing fine-grained details like finger motions remains a challenge, our model demonstrates better generation accuracy compared to existing methods. 
More qualitative results and video samples can be found in Appendix~\ref{sec:case_study_slg}.

At semantic level, our model also delivers higher BLEU-4 score, while for human evaluation, as shown in Table~\ref{tab:human_eval_result}, we report the average scores across five questions in the questionnaire~\ref{sec:questionaire} from ASL and CSL signers, respectively, on How2Sign and CSL-Daily. A higher average score means better quality. The poses generated by SignFLIP receive average scores of 1.69 and 1.94 on How2Sign and CSL-Daily datasets, respectively, while GT poses get scores of 3.48 and 3.44.
Notably, the ground-truth motions themselves also receive moderate ratings. This observation suggests an inherent limitation of 2D pose keypoints, which may not be sufficiently comprehensible for SL users. It also suggests the importance of finger movements in conveying precise meanings in SL communication. The generation of such fine-grained features remains a subject for future research.

\begin{figure}[t]
    \centering
  \includegraphics[width=\columnwidth]{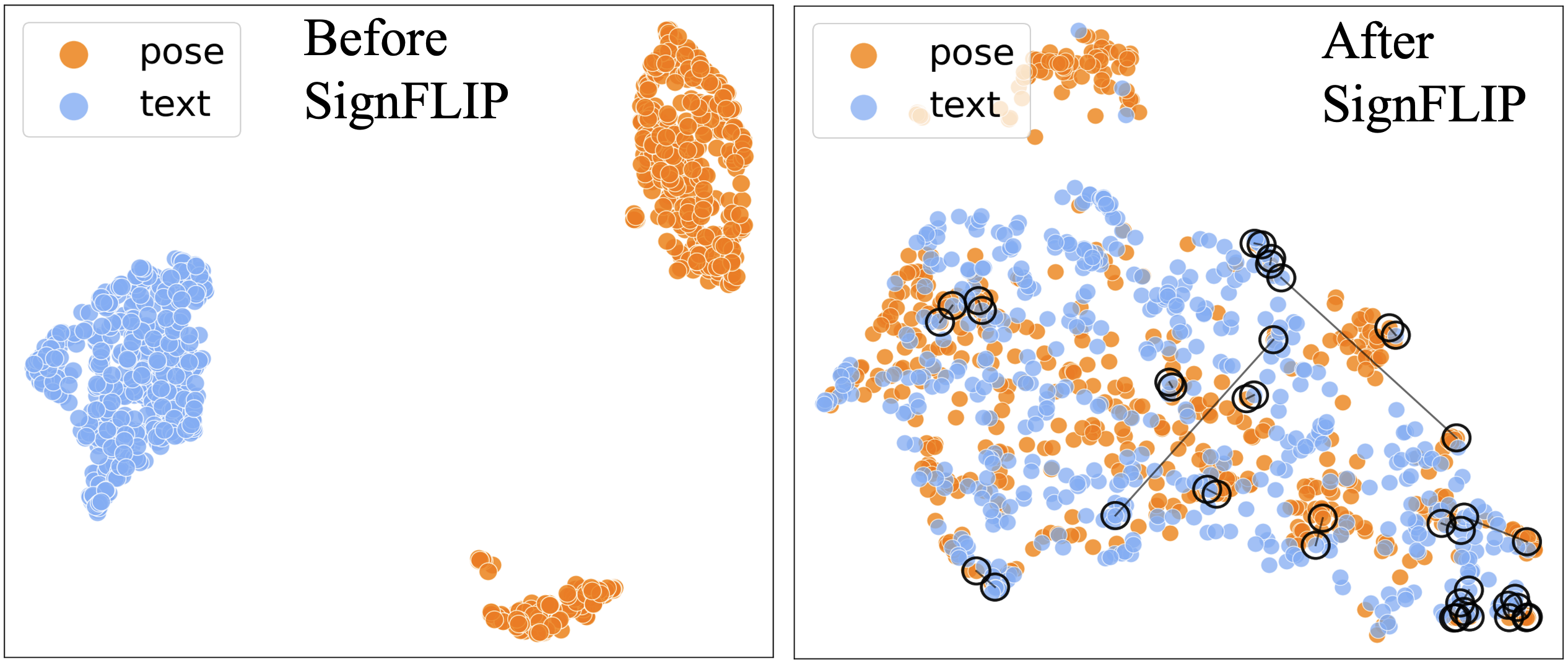}
  \caption{
  We visualize sign and text embeddings before and after SigFLIP, showing that initially separated feature distributions become closely aligned after training. Black-circled points and connecting lines indicate randomly selected sign--text pairs.
  }
  \label{fig:embeddings}

\end{figure}

\paragraph{Transferability to SLR.}
Unlike prior unified SL models that are typically trained from scratch and evaluated primarily on the tasks seen during training, SignFLIP builds on a general-purpose LLM and is pretrained on over 600K paired sign--text samples. As illustrated in Figure~\ref{fig:embeddings}, this pretraining encourages sign and text inputs to form a shared, semantically structured representation space. We therefore investigate whether the learned representations can transfer beyond SLT and SLG to Sign Language Recognition (SLR). Specifically, we consider both Continuous SLR (CSLR), which maps continuous sign videos to gloss sequences, and Isolated SLR (ISLR) on WLASL100, which classifies individual signs. We freeze the encoders in SignFLIP and attach task-specific decoders for the recognition tasks. As shown in Tables~\ref{tab:csldaily_cslr} and~\ref{tab:slr-qua}, SignFLIP achieves a test WER of 28.8 on CSL-Daily and a Top-1 accuracy of 88.87 P-I and 89.16 P-C on WLASL100. These results suggest that its pretrained representations capture transferable linguistic information and can serve as a versatile backbone for a broader range of SLP tasks.

\begin{table}[t]
\centering
\resizebox{0.48\textwidth}{!}{%
\begin{tabular}{l|cc|cc}
\toprule
\multirow{2}{*}{\textbf{Method}} 
& \multicolumn{2}{c|}{\textbf{Modality}} 
& \multicolumn{2}{c}{\textbf{WER$\downarrow$}} \\
\cmidrule(lr){2-3} \cmidrule(lr){4-5}
& \textbf{Pose} & \textbf{RGB} & \textbf{Dev} & \textbf{Test} \\
\midrule
SignBT~\citep{zhou2021improving}      &            & \checkmark & 33.2 & 33.2 \\
SEN~\citep{10.1609/aaai.v37i1.25164}  &            & \checkmark & 31.1 & 30.7 \\
CorrNet~\citep{hu2023continuous}      &            & \checkmark & 30.6 & 30.1 \\
MSLU~\citep{11125948}                 & \checkmark &            & 28.6 & 27.9 \\
Uni-Sign~\citep{li2025uni}            & \checkmark &            & 28.2 & 27.4 \\

\midrule
\textbf{SignFLIP (Frozen)}              & \checkmark &            & \colorbox{green!18}{\textbf{29.3}}& \colorbox{green!18}{\textbf{28.8}} \\
\bottomrule
\end{tabular}}
\caption{CSLR results on the CSL-Daily dataset with WER scores. SignFLIP demonstrates its strong transferability on CSLR task even with encoders frozen.}

\label{tab:csldaily_cslr}

\end{table}

\begin{table}[h]
\centering
\resizebox{0.5\textwidth}{!}{%
\begin{tabular}{l | cc | cc}
\toprule
\multirow{2}{*}{\textbf{Method}} & \multicolumn{2}{c|}{\textbf{Modality}} & \multicolumn{2}{c}{\textbf{WLASL100}} \\ 
\cmidrule(lr){2-3} \cmidrule(lr){4-5}
& Pose & RGB & P-I & P-C \\
\midrule
BEST~\cite{zhao2023best}       & \checkmark &            & 77.91 & 77.83 \\
SignBERT+~\cite{10109128}      & \checkmark &            & 79.84 & 80.72 \\
MSLU~\cite{11125948}           & \checkmark &            & 88.76 & 89.25 \\
NLA-SLR~\cite{zuo2023natural}  & \checkmark & \checkmark & 91.47 & 92.17 \\
\midrule
\textbf{SignFLIP (Frozen)}       & \checkmark &            & \colorbox{green!18}{\textbf{88.87}} & \colorbox{green!18}{\textbf{89.16}} \\
\bottomrule
\end{tabular}
}
\label{tab:wlasl100}
\caption{ISLR results on the WLASL100 dataset under the frozen setting. Without task-specific fine-tuning of SignFLIP encoders, the model achieves promising performance, demonstrating the transferability of its learned shared representations to downstream tasks.}
\label{tab:slr-qua}

\end{table}




\subsection{Ablation Study}

\paragraph{Stage-wise Alignment} is a crucial component of SignFLIP, as it balances the pretraining as well as two downstream tasks, while preventing too much degradation for SLT.
To examine its effect, we evaluate SignFLIP without pre-stages on translation and generation tasks.
For SLT, we conduct experiments on CSL-News and YouTube-ASL, two more challenging and noisier datasets, to better understand the performance limits of SignFLIP and to assess the extent to which pretraining benefits more difficult settings. While for SLG, we prove the importance of pretraining and SLT training by showing how much performance gain is brought by them individually on CSL-Daily dataset.

As shown in Table~\ref{tab:ablation_stage1_slt}, the setting with pretraining delivers quality gains of $+$ 3.37 BLEU-4 on CSL-News and $+$ 4.26 BLEU-4 on YouTube ASL. 
%
%
These findings demonstrate that the proposed pretraining effectively benefits subsequent downstream tasks.
The larger gains on challenging datasets further indicate that the pretraining helps the model better address the cross-modal gap between sign and text, enabling more robust learning under noisy and difficult conditions.


\begin{table}[h]
    \centering
    \scalebox{0.9}{
    \begin{tabular}{lcc}
    \toprule
    \multirow{2}{*}{\textbf{Setting}} & \textbf{CSL-News} & \textbf{YouTube-ASL} \\
                                      & \textbf{BLEU-4}    & \textbf{BLEU-4} \\
    \midrule
    w/o Pretraining& 18.24 & 4.17\\
    SignFLIP  & \textbf{21.61} & \textbf{8.43} \\
    \bottomrule
    \end{tabular}
    }
    \caption{Ablation study on stage design for SLT task.}
    \label{tab:ablation_stage1_slt}
\end{table}

The results of generation task are delivered in Table~\ref{tab:ablation_stage1_slg}. Pretraining and SLT training improved DTW performance by 11\% and 16\%, respectively. This demonstrates that the design generates high-quality joint sign-text representations from previous stages, enhancing SLG accuracy.


\begin{table}[h]
    \centering
    \scalebox{0.9}{
    \begin{tabular}{lcc}
    \toprule
    \textbf{Setting} & \textbf{DTW} & \textbf{DTW-MJE} \\
    \midrule
    w/o Pretraining \& SLT & 0.034 & 8.44e-5 \\
    w/o SLT & 0.030 & 7.37e-5 \\
    SignFLIP  & \textbf{0.025} & \textbf{6.15e-5} \\
    \bottomrule
    \end{tabular}
    }
    \caption{Ablation study on stage design for SLG task.}
    \label{tab:ablation_stage1_slg}
     \vspace*{-5mm}
\end{table}

\paragraph{Choice of LLM.} Recently, decoder-only LLMs have become the dominant architecture in generative tasks such as text and video generation, demonstrating remarkable performance. To investigate their potential in sign language, we conducted experiments to evaluate the performance of a widely used decoder-only LLM on SLT tasks. The results are summarized in Table~\ref{tab:ablation_llm_choices}. To ensure a fair comparison, all results in the table were obtained by training exclusively on the CSL-News dataset for SLT while only varying the LLM backbone. Specifically,  SignFLIP did not undergo pretraining.

\begin{table}[!htb]
    \centering
    \scalebox{0.85}{
    \begin{tabular}{lcc}
    \toprule
    \multirow{2}{*}{\textbf{Setting}} & \textbf{CSL-News} & \textbf{CSL-News} \\
                                      & \textbf{BLEU-4}    & \textbf{ROUGE} \\
    \midrule
    Qwen3~\cite{yang2025qwen3} & 5.19 & 20.07 \\
    mT5 (dec only) & 3.10 & 19.61 \\
    mT5 (enc-dec, w/o stage1)  & \textbf{18.24} & \textbf{39.13} \\
    \bottomrule
    \end{tabular}
    }
    \caption{Ablation study of LLM choices for SLT on CSL-News dataset.}
    \label{tab:ablation_llm_choices}
\end{table}

The interesting finding is that the performance of the decoder-only LLM does not outperform mT5-based settings, which aligns with the observations in SpaMo~\cite{hwang-etal-2025-efficient}.
These results suggest that, despite decoder-only models' powerful general capabilities, a significant gap remains between the data distributions and prompt formats of pretrained decoder-only LLMs and those of sign language data. For more capable models such as the Qwen series, this discrepancy may manifest as a stronger "\textit{Resistance to Alignment}"~\cite{ji2025language}, hindering the models from reaching their full potential. This implies that applying powerful decoder-only LLMs to sign language tasks still requires specialized designs to counteract this resistance while adapting their abilities and knowledge for sign language processing.
Ablation studies on SLG losses and the masking window size used during pretraining are provided in the Appendix.

\section{Conclusion}
In this paper, we present SignFLIP, a large-scale unified framework for sign language translation and generation that leverages a pretrained LLM and is trained on 600$\sim$700K sign--text samples. 
The experimental results demonstrate that the stage-wise training enhances translation and generation performance, delivering competitive results on several benchmarks.
Beyond the original tasks, SignFLIP also demonstrates promising transferability to downstream recognition tasks, including continuous and isolated sign language recognition.
These results suggest that SignFLIP learns generalizable sign representations and has potential as a foundation model for sign language processing.
We believe SignFLIP will inspire future research toward more scalable and general-purpose sign language models.


\section*{Limitations}

Our work has several limitations. First, most existing sign language datasets are not primarily based on Deaf and hard-of-hearing signers. For example, CSL-Daily is produced by sign language interpreters, meaning that the text serves as the original source. This limits its applicability to real-world scenarios and makes it less representative of real-world sign language use. Second, in our human evaluation, even the ground-truth samples received relatively low linguistic accuracy scores, suggesting that 2D pose representations may not be sufficiently understandable for sign language users. To enable more effective and accessible pose generation, we plan to explore the use of 3D avatars in future work. Finally, although sign language generation performance has improved, the generated hand movements still exhibit limited amplitude and tend to collapse toward mean poses. This may be partly attributed to the LLM itself, which can bias the outputs toward overly smooth and averaged motion patterns.

\section*{Acknowledgments}

This work was supported by KAKENHI 26H02525, ROIS NII Open Collaborative Research 261S04-24183, JST BOOST (Japan Grant Number JPMJBS2430) and the Science Tokyo Support Program for Doctoral Students, funded by the Universities for International Research Excellence. This study was carried out using the TSUBAME4.0 supercomputer at Institute of Science Tokyo. We also thank the anonymous reviewers for their constructive feedback, which improved the quality of this work.

\bibliography{custom}

\appendix

\section{Training Recipe of Each Stage}

\label{sec:training_recipe}
Table~\ref{sec:training_recipe} presents the recipes of each training stage.

\begin{table}[h]
\centering

\resizebox{0.9\columnwidth}{!}{%
\begin{tabular}{l|ccc}
\toprule
\textbf{Config} & \textbf{Stage 1} & \textbf{Stage 2} & \textbf{Stage 3} \\
\midrule
batch size             & 256 & 128 & 64 \\
optimizer              & \multicolumn{3}{c}{AdamW} \\
learning rate     & \multicolumn{3}{c}{$3\times10^{-4}$} \\
weight decay           & \multicolumn{3}{c}{$1\times10^{-4}$} \\
optimizer momentum     & \multicolumn{3}{c}{$\beta_1, \beta_2 = 0.9, 0.999$} \\
learning rate schedule & \multicolumn{3}{c}{cosine decay} \\
training epochs        & 20 & 50 & 50 \\

\bottomrule
\end{tabular}%
}
\caption{Training recipe of each stage.}
\label{tab:training_recipe}
\end{table}

While the hyperparameters that we adopt for loss weights are: $\alpha=0.05$, $\beta=1.0$, $\gamma=0.5$

\section{Statistics of Experimental Datasets}
\label{sec:selected_datasets}
Table~\ref{tab:selected_datasets} summarizes the statistics of the experimental datasets used in this work, including the language, sentence sample numbers, vocabulary size, recording hours, and data source.
\begin{table}[h]
\centering
\small
\setlength{\tabcolsep}{3pt}
\resizebox{\columnwidth}{!}{%
\begin{tabular}{@{}lccccl@{}}
\toprule
Name & Language & Samples & Vocab. & Hours & Source \\
\midrule
YouTube-ASL~\cite{10.5555/3666122.3667386} & ASL & $\sim$610K & 60K & 984   & Web \\
How2Sign~\cite{Duarte_CVPR2021}            & ASL & 33K & 16K & 79    & Lab \\
OpenASL~\cite{shi-etal-2022-open}          & ASL & 97K & 33K & 288   & Web \\
WLASL~\cite{li2020word} & ASL & 2,038 & 100 & Web\\
CSL-News~\cite{li2025uni}                  & CSL & $\sim$700K & 5K  & 1,985 & TV \\
CSL-Daily~\cite{zhou2021improving}         & CSL & 20K & 2K  & 23    & Lab \\
\bottomrule
\end{tabular}%
}
\caption{Selected ASL and CSL datasets.}
\label{tab:selected_datasets}
\end{table}
\section{SLT result on OpenASL}
\label{sec:openasl}
Table~\ref{tab:openasl} presents the performance of SignFLIP on OpenASL after Stage-2 fine-tuning.
\begin{table*}[t]
\centering
\small

\label{tab:openasl_results}
\resizebox{0.7\textwidth}{!}{%
\begin{tabular}{lccccc}
\toprule
\textbf{Method} & \textbf{Gloss} & \textbf{Visual} & \textbf{BLEU-4} & \textbf{ROUGE-L} & \textbf{BLEURT} \\
\midrule

\rowcolor{gray!20}
\multicolumn{6}{c}{\textbf{OpenASL}} \\

GloFE-VN~\cite{lin-etal-2023-gloss}         & \cmark &        & 7.06  & 21.75 & 36.35 \\
Conv-GRU$^{\dagger}$~\cite{8578910}         &        & \cmark & 4.58  & 16.10 & 25.65 \\
I3D-transformer~\cite{shi-etal-2022-open}   &        & \cmark & 5.66  & 18.64 & 28.82 \\
OpenASL~\cite{shi-etal-2022-open}           &        & \cmark & 8.59  & 21.02 & 31.09 \\
C$^2$RL~\cite{Chen2024C2RLCA}               &        & \cmark & 13.21 & 31.36 & --    \\

ShuBERT~\cite{gueuwou-etal-2025-shubert}    & \cmark & \cmark &
\colorbox{cyan!18}{\textbf{23.20}} &
-- &
\colorbox{cyan!18}{\textbf{60.60}} \\

Uni-Sign~\cite{li2025uni}                   & \cmark &        & 22.67 & 42.77 & 60.08 \\
Uni-Sign~\cite{li2025uni}                   & \cmark & \cmark & 23.14 &
\colorbox{cyan!18}{\textbf{43.22}} &
60.40 \\

\midrule
\textbf{SignFLIP (Ours)}                    & \cmark &        &
\colorbox{green!18}{\textbf{19.77}} &
\colorbox{green!18}{\textbf{41.56}} &
\colorbox{green!18}{\textbf{59.35}} \\

\bottomrule

\end{tabular}%
}

\caption{SignFLIP SLT performance on OpenASL after Stage-2 fine-tuning.}
\label{tab:openasl}
\end{table*}

\section{Ablation Study: MPM Window Size Impact}
In the pretraining stage, we randomly mask $n$ consecutive frames within a 24-frame window to help pose features capture semantic context. Table~\ref{tab:mpm-ablation} presents an ablation study on the impact of this window size.
\begin{table}[!htb]
\centering
\resizebox{0.4\textwidth}{!}{%
\begin{tabular}{lccc}
\toprule
\textbf{Setting} & \textbf{DTW} & \textbf{DTW-MJE} & \textbf{$\mathcal{L}_\text{align}$} \\
\midrule
16-frame  & 0.30 & 6.1e--4 & 0.17 \\
24-frame  & 0.12 & 2.7e--4 & 0.10 \\
32-frame  & 0.31 & 7.2e--4 & 0.15 \\
\bottomrule
\end{tabular}%
}
\caption{Ablation study with MPM window size on CSL-News dataset.}
\label{tab:mpm-ablation}
\end{table}

Results show that a 24-frame MPM window achieves optimal loss convergence and reconstruction results. This size balances sign reconstruction with broader semantic context, while other sizes overemphasize one objective and lead to suboptimal performance.

\section{Ablation Study: Loss for SLG} 
We further performed ablation studies to investigate the impact of the feature distillation and hand losses used in the SLG training. Results are summarized in Table~\ref{tab:slg-ablation}.

From the results, we notice that $L_\text{distill}$ yields a more pronounced improvement in generation performance. This is likely because, although the hand loss is more intuitive, optimization solely at the output pose level is too coarse-grained for the overall objective, making it difficult for the model to converge. In contrast, feature-level distillation provides supervision from multiple perspectives, helping the model establish clearer optimization targets and focus specifically on refining the nar transformer.

\begin{table}[!htb]
\centering
\resizebox{0.4\textwidth}{!}{%
\begin{tabular}{lcc}
\toprule
\textbf{Setting} & \textbf{DTW} & \textbf{DTW-MJE} \\
\midrule
 w/o $\mathcal{L}_\text{hand}$ and $\mathcal{L}_\text{distill}$  & 0.044 & 1.19e--4 \\
 w/o $\mathcal{L}_\text{hand}$                    & 0.029 & 7.84e--5 \\
 w/o $\mathcal{L}_\text{distill}$                 & 0.037 & 1.02e--4 \\
\midrule

 Ours                              & 0.025 & 6.15e--5 \\
\bottomrule
\end{tabular}%
}
\caption{Ablation study with SLG losses on CSL-Daily dataset.}
\label{tab:slg-ablation}
\end{table}

\section{Case Study of Translation}
\label{sec:case study}
Case studies of translation outputs on CSL-Daily, How2Sign, and OpenASL are shown in Table~\ref{tab:case_study}. All examples are taken from the test sets. Sentences in brackets denote our approximate English translations.

\section{Case Study of Generation}
\label{sec:case_study_slg}
We provide more cases on CSL-Daily and How2Sign as shown in Figure~\ref{fig:slg-qua-app}. Readers may also refer to the Google Drive link for video samples: \url{https://drive.google.com/drive/folders/1pZrFySsy-vEgrD7hkMC2KfDoZlRyVdpc?usp=sharing}
\begin{figure*}[!htb]
\centering
\includegraphics[width=\textwidth]{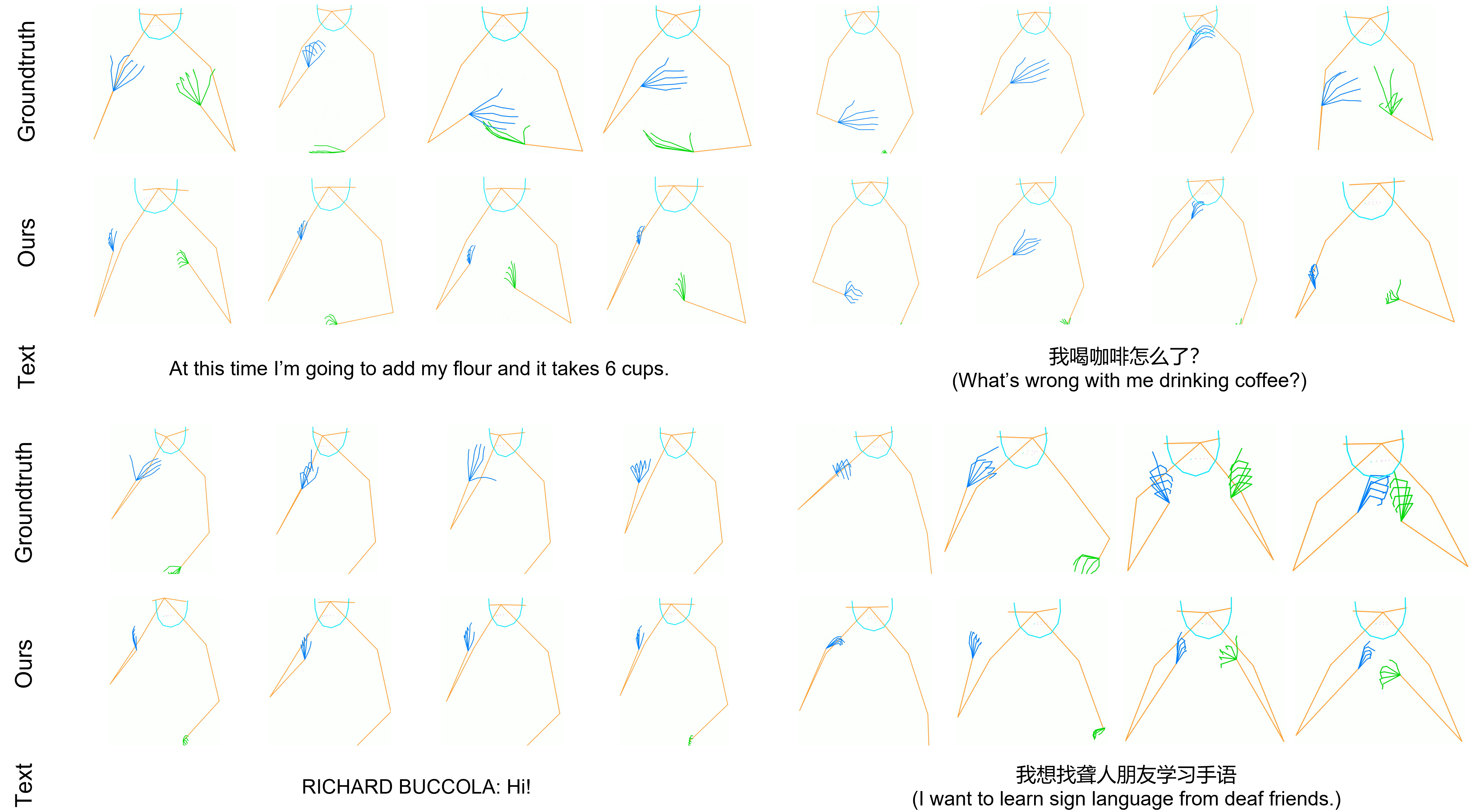} 
\caption{Qualitative results on both How2Sign (left) and CSL-Daily (right) datasets.}
\label{fig:slg-qua-app}
\end{figure*}

\begin{table*}[h]
\centering
\small
\resizebox{\textwidth}{!}{%
\begin{tabular}{ll}
\hline
\textbf{\textit{Systems}} & \textbf{Translation Output} \\
\hline

\multicolumn{2}{l}{\textbf{Examples from CSL-Daily}}\\
\multirow{2}{*}{\textit{Reference}} 
  & \begin{CJK*}{UTF8}{gbsn}你 和 小 张 什 么 时 候 认 识 的 ? \end{CJK*}\\
  & \textit{(When did you and Zhang meet?)}\\
\multirow{2}{*}{\textit{SignFLIP}} 
  & \begin{CJK*}{UTF8}{gbsn}你 的 小 张 什 么 时 候 认 识 的 ? \end{CJK*}\\
  & \textit{(When did you meet your Zhang?)}\\
\\

\multirow{2}{*}{\textit{Reference}} 
  & \begin{CJK*}{UTF8}{gbsn}我 不 去 爬 山 , 我 有 事 。\end{CJK*}\\
  & \textit{(I'm not going hiking; I have something to do.)}\\
\multirow{2}{*}{\textit{SignFLIP}} 
  & \begin{CJK*}{UTF8}{gbsn}我 不 去 爬 山 , 我 有 些 事 。\end{CJK*}\\
  & \textit{(I'm not going hiking; I have a few things to do.)}\\
\\

\multirow{2}{*}{\textit{Reference}} 
  & \begin{CJK*}{UTF8}{gbsn}你 喜 欢 这 条 裙 子 , 我 可 以 拿 一 件 给 你 试 穿 。\end{CJK*}\\
  & \textit{(If you like this dress, I can get one for you to try on.)}\\
\multirow{2}{*}{\textit{SignFLIP}} 
  & \begin{CJK*}{UTF8}{gbsn}你 喜 欢 穿 这 条 裙 子 , 我 可 以 给 你 穿 。\end{CJK*}\\
  & \textit{(If you like wearing this dress, I can help you put it on.)}\\

\hline
\multicolumn{2}{l}{\textbf{Examples from How2Sign}}\\
\textit{Reference} & I call it painting the wall.\\
\textit{SignFLIP} & And what this is, it's called painting the tool.\\
\\
\textit{Reference} & You need to be very careful when cleaning their cages if these birds are flighted.\\
\textit{SignFLIP} & So you have to be careful with that, because you don't have to worry about another thing.\\

\hline
\multicolumn{2}{l}{\textbf{Examples from OpenASL}}\\
\textit{Reference} & I've been staying home in self-quarantine and practicing social distancing.\\
\textit{SignFLIP} & For the past few years, we have been staying home, self-quarantining within ourselves and social distancing.\\
\\
\textit{Reference} & I transferred to Ohio state school and stayed there for two years.\\
\textit{SignFLIP} & I transferred to Ohio for educational school two years ago.\\

\hline
\end{tabular}%
}
\caption{Case study of translation outputs on CSL-Daily, How2Sign, and OpenASL. Examples are taken from the test sets. Sentences in brackets are our approximate English translations.}
\label{tab:case_study}
\end{table*}

\section{Human Evaluation Questionnaire}
\label{sec:questionaire}
We provide the questionnaire used in our human evaluation. Participants were asked to assess generated sign pose sequences on a 1--5 Likert scale.

\paragraph{Evaluation criteria.}
The evaluation covered the following aspects:
(1) Semantic accuracy,
(2) Motion smoothness,
(3) Anatomical plausibility, and
(4) Hand clarity.

\paragraph{Scale definition.}
For all questions, a score of 1 indicates very poor quality, while a score of 5 indicates excellent quality.

\paragraph{Instructions to participants.}
Participants were asked to watch each generated sign sequence and rate it according to the criteria above.

\paragraph{Questionnaire items.}
\begin{itemize}
    \item Without knowing the original sentence, can you understand this sign language video? 
    \item Please use a few keywords or a sentence to write down what you think this video means.  
    \item Compared to the reference text, does the motion accurately convey the original meaning?
    \item Is the skeleton motion continuous?
    \item Does the skeleton posture follow human physical laws?
    \item Are the finger motions and hand shapes clear?  
\end{itemize}
Participants were instructed to watch each generated sign video without first seeing the reference text and then assess its comprehensibility on a five-point scale.

\section{Potential Risks}
While we envision the advancements in LLMs and VLMs fostering global inclusivity, current SLP research remains constrained by limitations in data diversity and modeling accuracy. Users should exercise caution regarding the inherent risks of current translation and generation models, as potential inaccuracies or misinterpretations could lead to unintended negative social impacts.

\end{document}